%% file: _main.tex
\pdfoutput=1

\documentclass[11pt]{article}

\usepackage{acl}

\usepackage{times}
\usepackage{latexsym}

\usepackage[T1]{fontenc}

\usepackage{graphicx}
\usepackage{multirow}
\usepackage{multicol}
\usepackage{enumitem}

\usepackage[utf8]{inputenc}
\usepackage{CJKutf8}
\usepackage{booktabs}
\usepackage[english,bidi=default]{babel}
\babelprovide{arabic}
\usepackage[LAE, T1]{fontenc}
\addto\extrasarabic{\fontencoding{LAE}\selectfont}
\addto\noextrasarabic{\fontencoding{T1}\selectfont}
\usepackage{microtype}

\usepackage{amsmath}
\DeclareMathOperator{\LC}{LC}
\newcommand{\paxqa}{\textsc{PaxQA}}
\newcommand{\hwa}{\textsubscript{HWA}}
\newcommand{\awa}{\textsubscript{AWA}}

\title{\paxqa: Generating Cross-lingual Question Answering Examples at Training Scale}

\author{Bryan Li \and Chris Callison-Burch\\
  University of Pennsylvania\\ Philadelphia, USA \\
  \texttt{\{bryanli,ccb\}@cis.upenn.edu}
  }

\begin{document}
\maketitle

\begin{abstract}
Existing question answering (QA) systems owe much of their success to large, high-quality training data. Such annotation efforts are costly, and the difficulty compounds in the cross-lingual setting.
Therefore, prior cross-lingual QA work has focused on releasing evaluation datasets, and then applying zero-shot methods as baselines. This work proposes a synthetic data generation method for cross-lingual QA which leverages indirect supervision from existing parallel corpora. Our method termed \paxqa\ (\underline{P}rojecting \underline{a}nnotations for cross-lingual (\underline{x}) QA) decomposes cross-lingual QA into two stages. First, we apply a question generation (QG) model to the English side. Second, we apply annotation projection to translate both the questions and answers. To better translate questions, we propose a novel use of lexically-constrained machine translation, in which constrained entities are extracted from the parallel bitexts. 

We apply \paxqa\ to generate cross-lingual QA examples in 4 languages (662K examples total), and perform human evaluation on a subset to create validation and test splits. We then show that models fine-tuned on these datasets outperform prior synthetic data generation models over several extractive QA datasets. The largest performance gains are for directions with non-English questions and English contexts.
Ablation studies show that our dataset generation method is relatively robust to noise from automatic word alignments, showing the sufficient quality of our generations. To facilitate follow-up work, we release our code and datasets.\footnote{\url{https://github.com/manestay/paxqa}} 

\end{abstract}
\input{0introduction}
\input{1task}
\input{2related}
\input{3method}
\input{4experiments}
\input{5results}
\input{6analysis}

\input{7conclusion}


\bibliography{anthology,custom}
\bibliographystyle{acl_natbib}

\newpage
\input{Aappendix}

\end{document}

%% file: 0introduction.tex
\section{Introduction}

\begin{figure*}[ht]
    \centering
    \includegraphics[width=\textwidth]{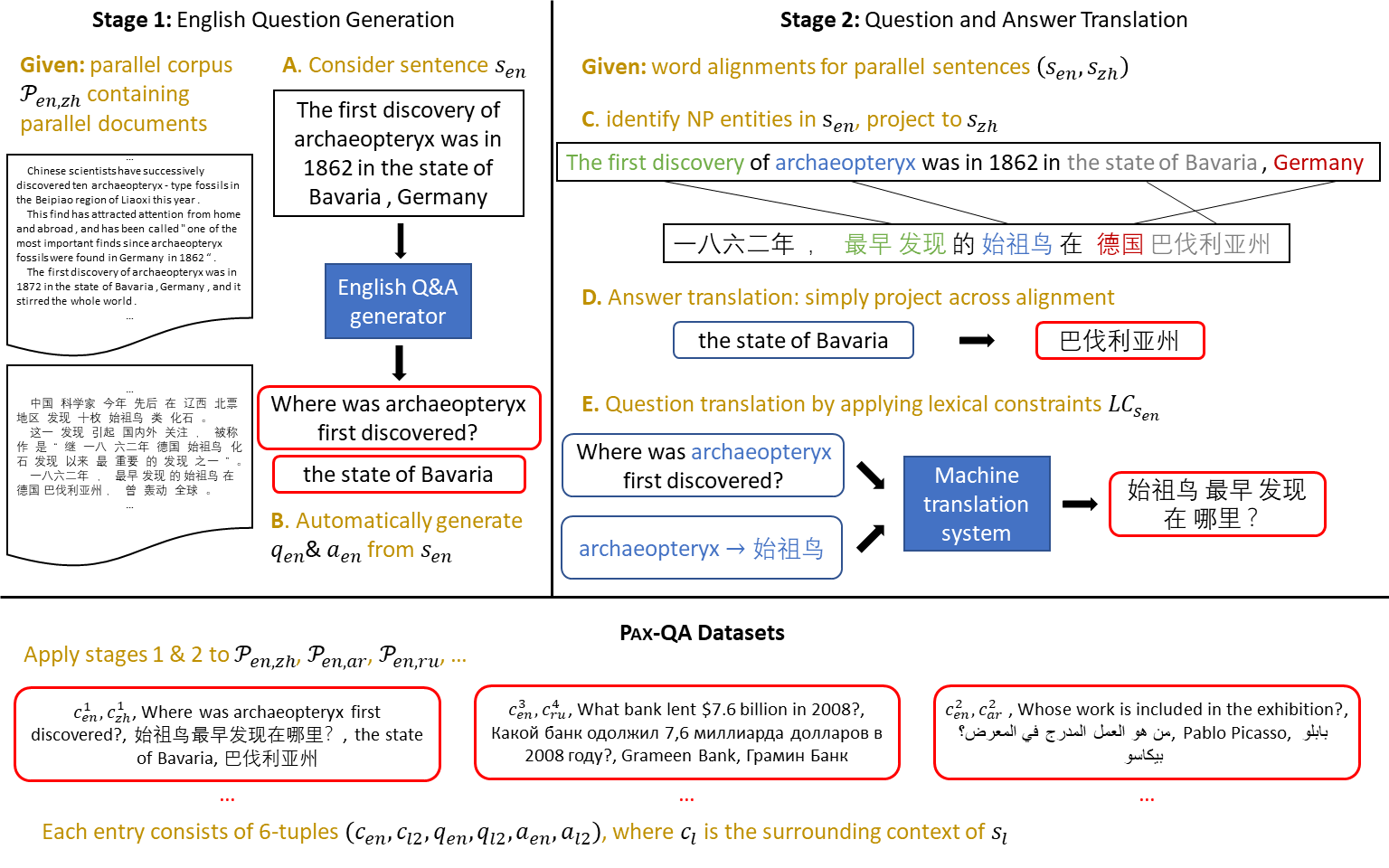}
    \caption{The \paxqa\ method generates a cross-lingual question-answering (QA) dataset given a word-aligned and parallel corpus. The two stages are English question generation (left), and question and answer translation (right). We run the pipeline on \{ar-en\}, \{zh-en\}, and \{ru-en\} datasets (bottom), resulting in 662K cross-lingual QA examples -- usable at training scale. Our \textit{generation} pipeline proceeds similarly to prior works' \textit{annotation} pipelines, but our method replaces all instances of human annotation with automated methods.}
    \label{fig:paxqa_pipeline}
\end{figure*}

A common framing of question answering (QA) in NLP is as a reading comprehension task, where questions about a specific text are to be answered by a span from a given context. Developing strong QA systems thus advances progress towards developing systems which can read and reason about texts. While earlier work developed QA models and resources in English only~\cite{rajpurkar2016squad,kwiatkowski2019natural}, recent work has sought to extend beyond English. Such datasets include \textsc{Ty}\textsc{Di} QA (11 languages; \citealt{clark2020tydi}), MLQA (6 languages; \citealt{lewis2020mlqa}), XQuAD (10 languages; \citealt{artetxe2020cross}), and MKQA (26 languages; \citealt{longpre-etal-2021-mkqa}), and are annotated with the help of native speakers of diverse languages. However, the high annotation cost required means they are limited to evaluation, and there is no data available for training.

These works therefore use several zero-shot approaches as baselines on their datasets. First, \textit{zero-shot transfer} involves fine-tuning a multilingual pre-trained language model (LM) on English QA data, then applying this model directly to multilingual QA. Another, \textit{translate-train}, uses a machine translation system (MT) to translate English data into other languages, then train new models on the translated data. Third, \textit{translate-test} instead uses an English QA model, at inference time translating other language QA to English, then back for the final evaluation. 

Alternatively, recent work has 
shown promising results with synthetic data augmentation~\cite{riabi-etal-2021-synthetic,shakeri-etal-2021-towards,agrawal2022qameleon}. In this approach, a question generation (QG) model is trained to generate synthetic multilingual QA examples, which are used as training data for a downstream QA model.

In this work, we propose a methodology for synthetic data augmentation, motivated by the insight that generating cross-lingual QA need not be entirely zero-shot. Instead, indirect supervision can be taken from widely-available parallel datasets originally collected for machine translation.
Our method termed \paxqa\ (\underline{P}rojecting \underline{a}nnotations for cross-lingual (\underline{x}) QA) extracts QA examples from parallel corpora. \paxqa\ decomposes the task into two stages: English question and answer generation, then machine translation of questions and answers informed by word alignments. Through this decomposition, \paxqa\ serves as a framework in which the individual QG and MT systems can be updated with the latest developments.

We apply our methodology to generate large-scale cross-lingual QA datasets in 4 languages: Chinese, Arabic, Russian, and English. We validate the quality of our generations by showing both improvements for downstream QA tasks, and by performing a human evaluation task. 

Our four key contributions are:

\begin{itemize}[noitemsep]
    \item We introduce \paxqa, a method to generate cross-lingual question answering (QA) datasets at training scale. Our method, depicted in Figure~\ref{fig:paxqa_pipeline}, requires no new models to be  trained, and instead decomposes cross-lingual QG into two automatic stages: 1) English question generation, and 2) word alignment-informed machine translation.
    \item To improve machine translation of questions, we propose a novel use of lexically-constrained machine translation. The lexical constraints are induced from the parallel sentences, and applied to the generated questions.
    \item We apply our method to generate cross-lingual QA datasets totaling 662K QA examples. Additionally, we ask human annotators to evaluate selected QA examples on several dimensions of quality; this results in 1,724 QA examples for \paxqa\ validation and test sets.
    \item We use our generated datasets to fine-tune extractive cross-lingual QA models. Our models significantly outperform zero-shot baselines for the in-domain evaluations, and outperform prior data generation methods on benchmark datasets such as MLQA. We perform ablations to show the robustness of \paxqa\ datasets created under various levels of noise.
\end{itemize}

%% file: 1task.tex
\section{Task Definition}

In this work, we focus on \textit{cross-lingual extractive question answering}. A cross-lingual extractive QA dataset consists of QA entries, each of which contains a context $c_f$, an answer $a_f$, and a question $q_e$ (where $f$ and $e$ denote a source and a target language). 
The task is defined as follows: given $q_e$ and $c_f$, a model must output an $a_f$ which is extracted from $c_f$. 
Our goal is to both propose a method to synthetically generate such a dataset, and to train a model to solve the task. 

Following prior works' broad definition of cross-lingual extractive QA~\cite{lewis2020mlqa}, it is possible that the two languages are the same ($e=f$). The only restriction is that in order to ensure the QA task is extractive, the context and answer must be in the same language; the question may or may not be in the same language. 

In the literature, multilingual QA\footnote{From here on, we use `QA' to mean `extractive QA'.} most commonly refers to the setting where QA is monolingual, but in multiple languages~\cite{clark2020tydi}. While the dataset as a whole considers multiple languages $l_1^n=l_1,...l_n$., each entry on its own is monolingual -- $(c, q, a)_{l_i}$, where $l_i \in l_1^n$. A cross-lingual QA dataset, in contrast, includes both monolingual and multilingual entries.

\paragraph{Notation} We denote a language pair as $\{f$-$e\}$. A language pair covers 4 cross-lingual directions ($f,e$; $e,f$; $f,f$; $e,e$). For each direction, the elements represent the language of the context and of the question, respectively. For example, $f,e$ means a context in $f$, and a question in $e$.

%% file: 2related.tex
\section{Related Work}
\subsection{Cross-lingual QA Resources}
Our work draws on two cross-lingual QA benchmarks: MLQA~\cite{lewis2020mlqa} and \textsc{XorQA}~\cite{asai-etal-2021-xor}. MLQA is multi-way parallel across 7 languages (49 directions total), and consists of 46k examples. \textsc{XorQA} is covers 7 languages, with each parallel to English (14 directions total), and consists of 40k examples. While the main dataset is open-domain, we use the extractive QA version, \textsc{XorQA}\textsubscript{GoldP}.

\subsection{Cross-lingual Question Generation}
Cross-lingual question-only generation was explored by~\citet{kumar-etal-2019-cross} and \citet{chiCrossLingualNaturalLanguage2020}. As answers are not provided, evaluation on QA tasks cannot be performed. These studies evaluate generated question quality both by human evaluations and by automated evaluations such as BLEU.

Concurrent to our work, \textsc{QAmeleon}~\cite{agrawal2022qameleon} generates synthetic multilingual QA datasets using prompt-tuning of a pretrained large language model. They find that these models can generate good quality QA using only five QA examples per language. While they only evaluate on multilingual QA datasets, it is likely that their method can easily be adapted to cross-lingual QA.

\citet{riabi-etal-2021-synthetic} and \citet{shakeri-etal-2021-towards} are most related to our work, as they also consider cross-lingual question and answer generation (QA generation, or simply QG). Both approaches adopt the view that as QG is the dual of QA, one can flip existing QA datasets into QG datasets. A supervised model can then be trained to generate cross-lingual QA examples, which are used as synthetic training data for a downstream QA model. \citet{riabi-etal-2021-synthetic} train their QG model on both English and machine-translated SQuAD data, and then train a separate QA model. \citet{shakeri-etal-2021-towards} train  a single model to perform both QA and QG. Their multitask setup consists of a QG task using SQuAD examples, and a masked language modeling task on \textsc{Ty}\textsc{Di} QA (without contexts).

The primary difference of our work versus these prior synthetic data generation approaches is that while they train custom models for cross-lingual QA generation, our work instead introduces a methodology that combines existing English QG systems and MT systems. Our approach is therefore more adaptable, as each component can be updated with state-of-the-art models as they are developed. Furthermore, a weakness of these prior approaches is that to train their models, they require supervised QA data for the languages of interest. which scarcely exist for low-resource languages. In contrast, our method only relies on parallel datasets with English, which are available even for low-resource languages.

\subsection{Annotation Projection}
Annotation projection is a time-tested technique which serves to transfer annotations from text in one language to parallel text in another language~\cite{li2022word}. It relies on word alignments, which can be learned in an unsupervised manner. In low-resource scenarios, annotation projection has been shown to be successful in cross-lingual NLP tasks, from parsing~\cite{hwa2005bootstrap, rasooli2017cross} to semantic role labeling~\cite{aminian-etal-2019-cross}. In this work, we apply annotation projection as a way to translate spans for QA: 
either the answer spans or the entities found in questions.

%% file: 3method.tex
\section{Data Generation Method}
We now describe the \paxqa\ data generation methodology, which was outlined in Figure~\ref{fig:paxqa_pipeline}. The goal is to generate synthetic cross-lingual QA datasets $\hat{\mathcal{D}}_{en,l}$, for each language of interest $l$ (\,$\hat{}$ specifies that it is generated). Our method assumes the availability of a parallel corpus $\mathcal{P}_{en,l}$ between English (en) and each $l$. We denote the two sides of a parallel corpus as $\mathcal{P}_{en}$ and $\mathcal{P}_{l}$. We also assume that these parallel corpora are word-aligned, either by humans or by a word alignment tool.
We now describe the two stages of our methodology.


\subsection{Question and Answer Generation}
In the first stage, we generate Q\&A pairs from $\mathcal{P}_{en}$. For this purpose, we adopt the pretrained question generation ($QG$) model from \citet{dugan-etal-2022-feasibility}.
They fine-tune T5 on three tasks: answer extraction, question generation and question answering. The intuition is that this multi-task learning setup improves individual task performance. We apply $QG$ directly to each sentence $s$ in $\mathcal{P}_{en}$. We then manually inspect several generated English Q\&A pairs, and implement heuristic filters to remove low-quality generations. For example, we filter answers containing question marks; others are listed in Appendix~\ref{sec:filters}. 

\paragraph{Using Paragraph Contexts}
Given that each question can be answered from the sentence $s$ it was generated from, we can further the challenge by setting the context $c$ to the entire paragraph where $s$ appears.
This modification also aligns \paxqa\ entries to SQuAD-style paragraph contexts. We therefore choose parallel corpora which have paragraph annotations. Note that using paragraph contexts is an optional post-processing step, and not a necessary part of the \paxqa\ method. \footnote{
To loosen this restriction, one can instead choose parallel corpora with article-level annotations, in which case a `paragraph' is simply $s$ + $N$ surrounding sentences. } 

\subsection{Question and Answer Translation}
In the second stage, we extend $\hat{\mathcal{D}}_{en}$ to the cross-lingual setting by translating both the question and the answer. We propose to use a translation process that is informed by word alignments, and which differs for questions and answers. 

To illustrate the process, let us consider a single generation from the prior stage, $q_{en}$ and $a_{en}$. These are generated from $s_{en}$, which consists of tokens $s_{en}^i$; we are also given that $s_{en}$ is parallel to $s_{l}$. A word alignment $wa$ links tokens from $s_{en}$ and $s_l$ which are translations of each other.

\subsubsection{Answer Translation} 
As $a_{en}$ is extracted from $s_{en}$, it corresponds to a set of tokens $s_{en}^u, s_{en}^w, ...$ with indices $u, w, ...$.  By applying the word alignment $wa$ on these indices, we translate to $l$ and obtain $a_{l} = s_{l}^x, s_{l}^y, ...$. Thus, answer translations come for ``free'' with the word alignments -- no MT system is needed.

Of course, the quality of this translation process depends on the quality of the word alignment. Word alignment errors will propagate to the projected answers.\footnote{Automated word aligners generally favor recall over precision. Therefore, alignment error are more likely to results in missing links, rather than incorrect ones.}  Furthermore, $wa$ may be under-specified, in which case not all (or none) of the words in $a_{en}$ can be projected. We discard those Q\&A pairs where either $a_{en}$ or $a_{l}$ are blank.

\subsubsection{Question Translation}
The translation of $q_{en}$ requires a machine translation (MT) system. In our work, we propose to use a lexically constrained MT system to that enforce lexical constraints within the text of questions. The rationale is that, since the task is already extractive with respect to answers, adding lexical constraints makes the questions more extractive as well. Furthermore, this can ensure the proper translation of more difficult terms, such as named entities, which are likely to be emphasized when evaluating reading comprehension. We perform question translation using three methods:

\begin{itemize}[noitemsep]
    \item For standard NMT, we use Transformer models~\cite{vaswani2017attention} (\textbf{vanilla}). We train a model for each language pair with data from WMT (see Appendix~\ref{sec:app_data}).
    \item We use the publicly available API for Google Translate~\cite{wu2016google} (\textbf{GT}). This is a strong translation system; however, it is not reproducible given its regular API updates. Also, the underlying mechanisms or training data are not specified.
    \item We also use a lexically constrained NMT system (\textbf{lex cons}), described below.
\end{itemize}




\paragraph{Lexically Constrained MT}
Lexically constrained MT adds to the model input a set of constraints $\LC$ which specify how specific source phrases should be translated into target phrases. In our work, we utilize the template-based MT model of \citet{wang2022template}.
The architecture is identical to a vanilla Transformer, but the method  modifies the data format  in order to incorporate the constraints as a template.

In prior work, constraints were sampled by choosing source phrases which were arbitrary sequences from $s_l$~\cite{chen2021lexically,post2018fast}. Our setting differs in that we apply these lexical constraints to the generated abstractive question $q_{en}$, instead of the context $s_{en}$ itself. Therefore, randomly sampling likely results in spans that do not appear in $q_{en}$. Using the insight that noun phrases (NPs) are more likely to be kept in $q_{en}$, we propose the key modification of sampling constraints by extracting NPs from $s_{en}$. We then keep only those constraints $\LC$ which appear in $q_{en}$.

\begin{CJK*}{UTF8}{gbsn}
An example is shown in Stage 2 of Figure~\ref{fig:paxqa_pipeline}. Of the three NPs in $s_{en}$, only `archaeopteryx' exists in $q_{en}$. So $LC = \{ \text{archaeopteryx} \rightarrow 
\text{始祖鸟}\}$. By contrast, random sampling may miss that constraint, and could output a span which does not appear in $q_{en}$, such as `1862 in the'.
\end{CJK*}

We retrain and reproduce the MT model of \citet{wang2022template} for our target languages. Then, we apply this NP constraint extraction process at inference-time. Compared to random sampling, our process allows for three times more lexical constraints to be used per question, on average.

%% file: 4experiments.tex
\section{Experimental Setup}

Although the \paxqa\ method can be applied to any language $l$ with English-$l$ parallel data, in our experiments we address three languages: Chinese (zh), Arabic (ar), and Russian (ru). We hope that the diversity in scripts and language families will illustrate the wide applicability of our method.

\subsection{Datasets Used}
Our proposed cross-lingual QA generation method requires both parallel corpora and word alignments. We provide further details in Appendix~\ref{sec:app_data}.

\paragraph{Parallel Corpora}
The machine translation community has made many parallel datasets publicly available.\footnote{Many corpora are at \url{https://opus.nlpl.eu}}
We use the News-Commentary (NC) and GlobalVoices (GV) datasets, which are multi-way parallel between many languages. We consider subsets which include 
English and our target languages. We also use the Arabic and Chinese GALE datasets from the LDC.\footnote{\url{https://catalog.ldc.upenn.edu}}

\paragraph{Word Alignments}
The GALE datasets include word alignments annotated by humans. 
For NC and GV datasets, we obtain alignments with the \texttt{awesome-align}~\cite{dou-neubig-2021-word} package. \texttt{awesome-align} induces alignments from a given multilingual LM. For zh, we use the provided fine-tuned checkpoint; for ru and ar, we use mBERT~\cite{bert}.

Parallel dataset statistics, as well as the number of Q\&A pairs generated from each dataset, are given in Appendix Table~\ref{tab:datasets}.

\subsection{\paxqa\ Datasets}
\label{ssec:paxqa_data}

We run our data generation method on each of the parallel corpora to obtain cross-lingual QA entries.

We designate the dataset created by concatenating QA entries generated from human word aligned datasets as \paxqa\hwa; likewise, from automatic word aligned datasets as \paxqa\awa. We further split into train/development/test sets; dev and test sets are created through human annotation filtering (further described in Section~\ref{ssec:eval_quality}). As shown in Table~\ref{tab:paxqa_data}, there are 578K cross-lingual QA entries for \paxqa\awa, and 82K for \paxqa\hwa.

Each \paxqa\ entry is a 6-tuple $(c_f, c_e, q_f, q_e, a_f, a_e)$. Recall that a language pair \{f-e\} consists of 4 cross-lingual QA directions. So training a model on \paxqa\hwa, for example, uses $82\text{K} * 4 = 328\text{K}$ training instances.

\begin{table}[t]
\centering
 \small
 \setlength{\tabcolsep}{4pt}
\begin{tabular}{@{}l|lll|lll@{}}
\toprule
 & \multicolumn{3}{c|}{\paxqa\awa} & \multicolumn{3}{c}{\paxqa\hwa} \\ \midrule
Lang & Train & Dev & Test & Train & Dev & Test \\ \midrule
zh & 72895 & 125 & 92 & 30963 & 104 & 190 \\ \midrule 
ar & 134912 & 193 &  87 & 51084 & 132 & 181 \\ \midrule
ru & 370534 & 300 & 320 & -- & -- & -- \\ \bottomrule
Total & 578341 & 618 & 499 & 82047 & 236 & 371  \\ \bottomrule
\end{tabular}
\caption{Number of generated cross-lingual QA entries per language $l$ and per split. Entries are cross-lingual between $l$ and English.
\paxqa\hwa is the dataset generated from the human word alignments, while \paxqa\awa is the dataset from automatic alignments.}
\label{tab:paxqa_data}
\end{table} 

%% file: 5results.tex
\section{Results}
\label{sec:results}

The \paxqa\ generation method creates synthetic QA data, we can then be used to fine-tune a cross-lingual extractive QA model. QA performance that beats prior work is our first way of validating the quality of the generations. We report results on three datasets: \paxqa\hwa\ test, MLQA~\cite{lewis2020mlqa}, and \textsc{XorQA}\textsubscript{GoldP}~\cite{asai-etal-2021-xor}.  The evaluation metric is the mean token F1 score, calculated with the official MLQA script.\footnote{\url{https://github.com/facebookresearch/MLQA/blob/main/mlqa_evaluation_v1.py}}


Prior work reports average results across all cross-lingual directions.
We instead group directions as follows: 
`non-en $q$'uestion + en context; 
`en $q$'uestion + non-en context; 
`monolingual', and `non-en (xling)' cross-lingual. We then analyze the results by group, which gives us a clearer picture of how the methods affect different directions. On MLQA, we discuss results only in \{ar,zh,en\}; results averaged over all 49 MLQA directions are in Appendix Table~\ref{tab:mlqa_all}. 

\paragraph{QA Model} We adopt XLM-R~\cite{conneau2020unsupervised} as our QA model, and initialize to the pretrained large checkpoint from the \texttt{transformers} library~\cite{wolf2019huggingface}.  Following the advice of~\cite{alberti-etal-2019-synthetic}, we fine-tune in two rounds: first on synthetic QA data, then on SquAD. We find that this improves results over shuffling real and synthetic data.

\paragraph{QA Data} 
We use the \paxqa\hwa\ and \paxqa\awa\ data splits.
Recall that each \paxqa\ entry gives us 4 cross-lingual entries ($f,e$; $e,f$; $f,f$; $e,e$). For training, we also use SQuAD; for validation, we also use MLQA and \textsc{XorQA}\textsubscript{GoldP}.

\subsection{QA Models for Comparison}

For transfer learning approaches, we consider only \textbf{zero-shot}, since the translate-test and translate-train baselines require significant computational overhead. Prior work compared these three baselines~\cite{lewis2020mlqa, longpre-etal-2021-mkqa} and found that zero-shot and translate-train perform similarly, while translate-test performs poorly because it uses needs to translate inference data twice: the input in $l$ to English, then the  prediction in English to $l$.

We compare \paxqa -trained models to the zero-shot baseline results reported by MLQA. We also compare to~\citet{riabi-etal-2021-synthetic}, whose results are directly comparable to ours because the same underlying XLM-R model is used.

\subsection{In-domain Results}

\begin{table}[t]
\centering
\small
\setlength{\tabcolsep}{3pt}
\begin{tabular}{@{}l|ll|ll|lll|l@{}}
\toprule
 &
  \multicolumn{2}{c|}{\textbf{non-en $q$}} &
  \multicolumn{2}{c|}{\textbf{en $q$}} &
  \multicolumn{3}{c|}{\textbf{monolingual}} \\ 
$q$ MT   & en,zh & en,ar & ar,en & zh,en & ar,ar & zh,zh & en,en & avg \\ \midrule
SQuAD      & 67.0  & 78.9  & 85.8  & 83.5  & 79.8  & 73.9  & 90.9 & 80.0 \\
vanilla  & \textbf{91.8}  & 90.6  & 87.7  & 88.4  & 87.8  & 82.2  & 92.7 & 88.7 \\
lex cons & 88.9  & 91.5  & \textbf{90.4}  & 86.8  & \textbf{90.2}  & 84.1  & \textbf{93.9} & 89.4 \\
GT       &  88.0  & \textbf{93.4}  & 89.7  & \textbf{89.0}  & 88.3  & \textbf{85.0}  & 93.3 & \textbf{89.5}  \\ 
\bottomrule
\end{tabular}
\caption{\paxqa\hwa\ test F1 scores for XLM-R models trained on various datasets. Rows 2-4 are trained on only \paxqa\hwa, differing in the question translation method. `$e,f$' indicates the context and answer are in $e$, while the question is in $f$.
}
\label{tab:paxqa_results}
\end{table}

Table~\ref{tab:paxqa_results} shows the results on the \paxqa\hwa\ test sets. All 4 models are initialized to XLM-R, then further fine-tuned on SQuAD
or on \paxqa\hwa. The scores of the three \paxqa\hwa-trained models 
are fairly close. This is likely because they use the same context and answers, and differ only in how English questions are translated. 
Still, we see that lex cons beats vanilla overall, and performs about the same as GT. This is notable because GT is a much stronger MT system than our bilingual Transformer-based models.

The largest improvements with training on \paxqa\ data are for `non-en $q$'. The lex cons model achieves +21.9 F1 for en,zh (88.9 > 67.0), and +12.6 F1 for en,ar (91.5 > 78.9). We also see significant improvements for `monolingual' (+10 F1), and modest ones for `en $q$' (+3-5 F1).

\subsection{Generalization Results}

\begin{table*}[ht]
\centering
\small
\setlength{\tabcolsep}{4pt}
\begin{tabular}{@{}llp{2.3cm}l|ll|ll|lll|ll|l@{}} \toprule
  & & & &
  \multicolumn{2}{c|}{\textbf{non-en $q$}} &
  \multicolumn{2}{c|}{\textbf{en $q$}} &
  \multicolumn{3}{c|}{\textbf{monolingual}} &
  \multicolumn{2}{c|}{\textbf{non-en xling}} & \\ 
& \# & Method & Train Data        & en,zh & en,ar & zh,en         & ar,en & ar,ar & zh,zh & en,en & ar,zh         & zh,ar & avg \\ \midrule
\multirow{4}{*}{\rotatebox{90}{baseline}} & 1 & zero-shot \cite{lewis2020mlqa} & SQuAD (S) & 53.9  & 50.8  & 52.9          & 60.0  & 54.8  & 61.1  & 74.9  & 43.5          & 40.9 & 54.8  \\
& 2 & zero-shot & SQuAD (S)          & 62.5  & 64.8  & 68.8          & 67.1  & 65.6  & 70.1  & 84.3  & 53.3          & 54.8 & 65.7 \\
& 3 & \citet{riabi-etal-2021-synthetic} & synthetic + S          & 77.8 & 76.3 & 66.8 & 68.9 & 66.6 & 67.7 & 83.9 & \textbf{64.9} & 61.7 & 70.5 \\ \midrule
\multirow{4}{*}{\rotatebox{90}{human WA}} & 4 & \paxqa\hwa & lex cons  & 72.6  & 70.2  & 64.6         & 61.1  & 60.1  & 65.1  & 77.9  & 57.2          & 58.3 & 65.2 \\
& 5 & \paxqa\hwa & vanilla + S  & 77.9 & 75.9 & 69.8 & 68.8 & \textbf{67.0} & 70.7 & 84.6 & 64.1 & 63.7  & 71.4  \\
& 6 & \paxqa\hwa & lex cons + S & \textbf{78.9} & \textbf{76.7}          & \textbf{70.2}          & \textbf{69.2}          & \textbf{67.0}          & \textbf{71.1}          & \textbf{84.8} & 64.3 & \textbf{64.9} & \textbf{71.9} \\
& 7 & \paxqa\hwa & GT + S     & 77.7 & 75.0 & 69.3 & 68.3 & 66.4 & 70.2 & 84.1 & 63.2 & 63.2 & 70.8  \\  \midrule
\multirow{2}{*}{\rotatebox[origin=c]{90}{\parbox[c]{.5cm}{\centering auto WA}}} & 8 & \paxqa\awa  & lex cons & 69.5 & 67.8 & 61.7 & 60.5 & 59.6 & 63.1 & 77.0 & 55.8 & 56.0 & 63.4 \\ 
& 9 & \paxqa\awa & lex cons + S & 77.3 & 74.6 & 69.8 & 69.0 & 67.0 & 70.3 & 84.5 & 63.4 & 62.5 & 70.9 \\
\bottomrule
\end{tabular}
\caption{MLQA test F1 scores for models trained on various datasets. The model in row 1 is XLM, and in the other rows is XLM-R. The \paxqa\ rows are obtained by training on generated cross-lingual QA pairs from parallel datasets, which are either human word-aligned (rows 3-7) or automatically word-aligned (rows 8-9). To translate questions from English, the systems are \underline{vanilla} NMT, \underline{lex}ically \underline{cons}trained NMT, or \underline{G}oogle \underline{T}ranslate. }
\label{tab:mlqa_results}
\end{table*}

We now evaluate how well \paxqa\hwa-trained models generalize to MLQA. These results can be considered out-of-domain, as MLQA was collected over Wikipedia, while \paxqa\ was generated from news articles. Results are shown in Table~\ref{tab:mlqa_results}.

Our best overall model is lex cons + SQuAD (row 6), though the other \paxqa\ variants perform similarly (rows 5-7).  Compared to the zero-shot baseline (row 2), our best model improves most significantly for `non-en $q$': +16.4 F1 for en,zh (78.9 > 62.5); +11.9 F1 for en,ar (76.7 > 64.8). `non-en xling' achieve +10 F1, while `en $q$' and monolingual performance are similar to the baseline. Our method outperforms the prior best synthetic generation method (row 3) by 1.4 F1 overall. 

We also observe that, as expected, a model trained on \paxqa\hwa\ alone (row 4) under-performs those with both \paxqa\hwa\ and SQuAD; the same goes for \paxqa\awa\ (row 8 vs. row 9).



\subsection{Results using Automatic Word Alignments}
\label{sec:awa_results}
\begin{table}[t]
\centering
\begin{tabular}{@{}ll|ll@{}}
\toprule
Method     & Train Data   & en,ar & en,ru \\ \midrule
zero-shot & SQuAD (S) & 57.4 & 71.2 \\
\paxqa\hwa & lex cons & 58.6 & 68.1 \\ 
\paxqa\hwa & lex cons + S & \textbf{68.9} & 71.1 \\ 
\paxqa\awa & lex cons + S & 64.7 & \textbf{72.3} \\ \bottomrule
\end{tabular}
\caption{\textsc{XorQA}\textsubscript{GoldP} test F1 scores for models trained on various datasets. All rows are based on a fine-tuned XLM-R model. }
\label{tab:xorqa_results}
\end{table}

In this section, we report \paxqa\awa\ results.\footnote{We also trained on a concatenation of \paxqa\hwa\ and \paxqa\awa\ data, but results were about the same.} In this setting, word alignments are noisy and include many alignment errors.

Results on MLQA are shown in Table~\ref{tab:mlqa_results}. Comparing the best \paxqa\hwa\ model (row 6) and the best \paxqa\awa\ model (row 9), we see performance only drops by 1.0 F1 across overall. Still, `non-en $q$'
and `non-en xling'  results of \paxqa\awa\ handily beat the baseline model (row 2), and monolingual F1 scores are similar.

We also report results for \textsc{XorQA}\textsubscript{GoldP} dev sets for en,ar and en,ru\footnote{\textsc{XorQA} does not release the test set answers, and ru,en and ar,en are not supported.}, as shown in Table~\ref{tab:xorqa_results}.  For en,ar \paxqa\hwa\ achieves +11.5 F1 (68.9 > 57.4). For en,ru, \paxqa\awa\ achieves +1.1 (72.3 > 71.2); \paxqa\hwa\ has not seen any en,ru synthetic data, so it performs the same as the baseline.


\paragraph{Discussion}
We find that the \paxqa\hwa -trained models perform the best, notably achieving a new state-of-the-art on MLQA over ~\citet{riabi-etal-2021-synthetic}. The significant but relatively small improvements (1-2 F1) in downstream QA performance are expected, as the underlying approaches (training on synthetic data, then real data) and models (fine-tune of XLM-R) are similar. Still, they likely reflect larger improvements in the upstream QG task. As \paxqa\awa-trained models perform only 1.0 F1 lower than \paxqa\hwa, 
our method is relatively robust to noise from automated alignments.\footnote{For example, alignment error rate of \{zh-en\} is 13.4.} Recalling that our method does not require non-English QA data, these characteristics show that \paxqa\ is effective and extensible beyond the 4 languages considered here.

%% file: 6analysis.tex
\section{Analysis}

\subsection{Evaluating the Quality of Generations}
\label{ssec:eval_quality}
We run a human annotation task to evaluate the quality of generations. At a high-level, we adapt the methodology of \citet{dugan-etal-2022-feasibility}.
We sample 2,921 QA entries from the \paxqa\ generations for the evaluation task. Of these, 1,724 (59.0\%) were acceptable to human annotators. We then randomly split these entries into development and test sets. More details are provided in Appendix~\ref{sec:heval_task}.

\subsection{Ablations}
\label{sec:ablation}

Prior work~\cite{shakeri-etal-2021-towards,riabi-etal-2021-synthetic} showed that a synthetic data training scheme allows for cross-lingual generalization. This means that even training with a single language pair improves cross-lingual QA performance for all directions, including unseen ones.  From this, they hypothesize that multilingual models such as XLM-R already possess good multilingual internal representations, and this scheme allows for generalization to non-English QA. We verify this hypothesis by performing several ablations on our datasets.

\paragraph{Bilingual QA Models}
Instead of training a single model to perform cross-lingual QA for all pairs, we can train bilingual models. We train one model on \paxqa\hwa\ {zh-en} and another on {ar-en} \paxqa\hwa. Results on MLQA are shown in Appendix Table~\ref{tab:app_mlqa_results}. We see that the multilingual model outperforms the bilingual models overall (71.9 > 70.7), most notably for the `non-en xling` directions. Interestingly, we observe that the {ar-en} model, which is zero-shot with respect to zh, outperforms {zh-en} for en,zh and zh,zh. Results on \paxqa\hwa\ test are shown in Appendix Table~\ref{tab:app_paxqa_results}. We again observe the multilingual model outperforms the bilingual models (89.4 > 87.9). These findings are evidence for the cross-lingual generalization hypothesis from above.

\paragraph{Extending to non-English Parallel Directions}
In our prior experiments, we generated English-centric QA entries. However, we can extend to non-English parallel directions by pivoting a multi-way parallel dataset through English. We apply the following pivoting strategy to the News-Commentary parallel corpora. We first take the articles which are parallel between 3 languages: zh, ar, and en.
The Q\&A generation stage of \paxqa\ remains the same, since it operates on English. In the Q\&A translation stage, we perform our alignment-informed translations from en to ar, from en to zh. We now have question, answers, and contexts in both languages, giving us an \{ar-zh\} cross-lingual extractive QA dataset. 

Results for this fine-tuned model (ar-zh pivot data + SQuAD + lex cons) 
are shown in Table~\ref{tab:app_mlqa_results}. F1 scores are overall slightly lower than prior models. This is even the case for the ar,zh and zh,ar directions. A possible reason is because the pivoting strategy compounds noise, since we apply automatic alignments twice, and perform MT twice.

\subsection{Case study: Comparing Question Translation Methods}
\label{ssec:case}

\begin{CJK*}{UTF8}{gbsn}
\begin{figure}[t]
    \centering
    \noindent\fbox{ \small 
        \parbox{0.95\linewidth}{
        \textbf{$c_{en}$}: The first discovery of \textcolor{blue}{archaeopteryx} was in 1862 in the state of Bavaria, Germany.\\
        \textbf{$q_{en}$}: Where was \textcolor{blue}{archaeopteryx} first discovered? \\
        $a_{en}$: the state of Bavaria, $a_{zh}$: 巴伐利亚 \\
        \textbf{$q_{zh}$ (vanilla)}: 最早的考古发现在哪里？ \\
        (en: Where was the earliest archaeological discovery?) \\
        \textbf{$q_{zh}$ (lex cons)}: \textcolor{blue}{始祖鸟}最早发现在哪里？ \\
        (en: Where was archaeopteryx first discovered?) \\
        \textbf{$q_{zh}$ (GT)}: 最早发现的最早的考古学是哪里？ \\
        (en: Where was the earliest discovery of the earliest archaeology?)
        }
    }
    \caption{An example entry from \paxqa\hwa. }
    \label{fig:case_study}
\end{figure}

In Section~\ref{sec:results}, showed that  models using 3 different methods of question translation performed similarly. As a case study, consider the example shown in Figure~\ref{fig:case_study}. Because ``archaeopteryx'' is an uncommon word, the vanilla and GT systems fail to translate it properly; it is instead translated incorrectly to 考古\ (archaeological) and 考古学\ (archaeology). The lexically constrained system gets it correct, because it has the constraint given as input.

Despite the incorrect question translations, it is easy to see how any NMT system could derive the correct answer by simply noticing that the question asks for ``where'' (in either language), and could just return the location ``the state of Bavaria''. This is a known issue with reading comprehension-style questions~\cite{kwiatkowski2019natural}.

\paragraph{A Harder Task} The case study and the relatively high QA results suggest that even a cross-lingual formulation of the extractive QA task is fairly easy.
We identify \textit{round-trip cross-lingual QA} as the immediate next step. For this task,  given $(c_f, q_e)$, a model must predict $a_e$.\footnote{\citet{asai-etal-2021-xor} explores the related round-trip task for open-domain cross-lingual QA.}  While the answer can still be found in the context, it must now be translated back to the question's language (i.e., round-trip). This would be more useful to end-users who would like to be able to ask questions of multilingual documents, and receive answers they can understand.
The \paxqa\hwa\ and \paxqa\awa\ datasets can indeed be used for this new task. However, the modeling approaches covered here do not support it, and we leave such efforts to future work.
\end{CJK*}

%% file: 7conclusion.tex
\section{Conclusion}

We present \paxqa, a synthetic data generation methodology for cross-lingual QA which leverages indirect supervision from parallel datasets. We decomposed the task into two stages: English QA generation, then QA translation informed by annotation projection. Unlike prior methods, \paxqa\ neither requires training of new models, nor non-English QA data. This means our method can even be applied to low-resource languages. We proposed the novel use of lexically-constrained MT to better translate questions, which assists in proper translation of uncommon entities. Finally, we showed that training on \paxqa\ data allows downstream models to significantly outperform baselines, and achieve a new zero-shot state-of-the-art on the MLQA benchmark. We close by reiterating that \paxqa\ is a methodology, and that as such, the machine components for MT and word alignment can be swapped out as new developments progress in those fields.

\section{Limitations}
The main limitation of our method is that it requires datasets which are parallel to English. However, because of the great efforts placed into collecting resources for machine translation, such datasets are reasonably available. In the MT field, ``low-resource'' generally means less than 1M parallel sentences~\cite{haddow-etal-2022-survey}. This is ample enough data to train automatic word aligners between English and some language, needed by our method. 

Because of resource constraints on our end, we only ran our method end-to-end for three languages. However, we have claimed that by decomposing cross-lingual QG into English QG and MT steps, our method allows for QA generation in low-resource languages. As an initial step, we are running the \paxqa\ pipeline on the FLoRes v1~\cite{guzman-etal-2019-flores} dataset, which covers Nepali and Sinhala. After the dataset is generated, we will investigate how we can evaluate the quality of generations for these languages not been studied by the QA community yet. While back-translation using NMT could be a first start, more likely this requires finding native human annotators.

Beyond the parallel dataset limitation, we acknowledge that the English-centric nature of our approach is not ideal. We inherit this problem from the general body of cross-lingual QA research. For most prior collected datasets, human annotators required English fluency -- to translate questions from English to their native language, and even to be able to read instructions written in English. Still, future research should aim to study all languages more equally. Our ultimate goal, as we discussed in Section~\ref{ssec:case}, is to develop QA models that allow users to pose questions regarding documents in any language, and receive an answer back in their native language. Given that the bulk of the information available on the web is in English, such a system would allow for more equitable access to the world's information resources for all humans.

Another set of limitations concern the quality of our question generations. For the off-the-shelf model we used, only 59.0\% of generations were deemed acceptable. The \paxqa\ approach allows for drop-in replacements of the English QG system, and follow-up work can use stronger QG systems, and therefore improve the final results. Also, our human evaluation task focused on the English side of cross-lingual QA entries. This is because our annotators were students in the USA, and therefore we did not expect them to be multilingual. We checked the quality of the translated answer through back-translation, but this is only a proxy. Furthermore, we did not perform human annotation on the non-English question.

\paragraph{Ethical Considerations} The human annotators we recruited were given  instructions, and could then contribute between 0–100 tasks. We believe that the extra credit for their final course grade was a fair incentive.

The synthetic data generation method we used can possibly generate misleading or even toxic information, depending on the contexts it is given. Some of our human annotators flagged certain generations for our review. The culprit was contexts which expressed someone's opinion; for example, an interview with a controversial politician. In such cases, the generated questions  were from the perspective of that person. From the 3K annotations we did, we discarded any QA entries that were deemed unacceptable. However, we do not verify all 600K+ examples we release. We do apply some filtering steps to attempt to mitigate low-quality generations. Furthermore, we have only run our QA generation method on news datasets which are widely used and understood within the general community.

\section{Acknowledgements}
This research is based upon work supported in part by the Air Force Research Laboratory (contract FA8750-23-C-0507), the DARPA KAIROS Program (contract FA8750-19-2-1004), the IARPA HIATUS Program (contract 2022-22072200005), and the NSF (Award 1928631). Approved for Public Release, Distribution Unlimited. The views and conclusions contained herein are those of the authors and should not be interpreted as necessarily representing the official policies, either expressed or implied, of AFRL, DARPA, IARPA, NSF, or the U.S. Government.

We would like to thank Marianna Apidianaki for her detailed feedback. We would also like to thank Yang Chen, Chao Jiang, Alyssa Hwang, and the anonymous reviewers for their input.

%% file: Aappendix.tex
\appendix
\section{Details on Datasets Used}
\label{sec:app_data}
The parallel corpora we use in this work come from three datasets: GALE, NewsCommentary, and GlobalVoices. The latter two come from OPUS~\cite{opus}. Dataset statistics are given in Table~\ref{tab:datasets}.

GALE\footnote{\url{https://catalog.ldc.upenn.edu/LDC_NUMBER}} are a collection of parallel news datasets available on LDC. These are word-aligned by trained human annotators. For Arabic-English, we use the following LDC numbers: LDC2013T10, LDC2013T14, LDC2014T03, LDC2014T05, LDC2014T08, LDC2014T10, LDC2014T14, LDC2014T19. For Chinese-English, we use: LDC2012T16, LDC2012T20, LDC2012T24, LDC2013T05, LDC2013T23, LDC2015T04, LDC2015T06.

GlobalVoices\footnote{\url{https://opus.nlpl.eu/GlobalVoices.php}} is a parallel corpus of news articles in 46 languages. We use only the ar-en and ru-en subsets of the data. While standard (zhs) and traditional Chinese (zht) are part of this corpus, we do not use them because we have found that the zh-en sentence alignments are of very poor quality. As the sentence alignments for other GlobalVoices directions are near perfect, we suspect some preprocessing issue occured.

News-Commentary\footnote{\url{https://opus.nlpl.eu/News-Commentary.php}} is a parallel corpus of news commentaries in 15 languages. We use only the ar-en, zh-en, and ru-en subsets of the data. Note that both GlobalVoices and News-Commentary parallel between almost all languages considered. Our work only generates questions (originally) in English, which requires the parallel with English restriction. We leave future work to generate questions directly from multiple languages.

\section{Filtering Lower-Quality Generations} 
\label{sec:filters}
We implement heuristic filters to remove any generations that have the following properties:
\begin{enumerate}[noitemsep]
    \item The generation is a duplicate.
    \item The question is of the form "What is the answer…".
    \item The answer contains a question mark.
    \item The source sentence is less than 5 tokens, not including punctuation.
    \item The answer (either $a_{en}$ or $a_{l}$) consists of only punctuation.
\end{enumerate}

We note that most of these issues can be addressed by a higher quality question generation system. We leave this to future work, and note that the \paxqa\ method is an orthogonal contribution to those developments.

\begin{table}[ht]
\centering
\begin{tabular}{@{}llll@{}}
\toprule
lang & dataset & \# sents & \# QA gen \\ \toprule
\multirow{2}{*}{zh} & NewsComm & 73623 & 73112 \\ \cmidrule(l){2-4} 
 & \textbf{GALE} & 31390 & 31257 \\ \midrule 
\multirow{3}{*}{ar} & NewsComm & 80119 & 80863 \\ \cmidrule(l){2-4} 
 & GlobalVoices & 58985 & 54329 \\ \cmidrule(l){2-4} 
 & \textbf{GALE} & 49568 & 51397 \\ \midrule
\multirow{2}{*}{ru} & NewsComm & 203598 & 208190 \\ \cmidrule(l){2-4} 
 & GlobalVoices & 164895 & 162964 \\ \bottomrule
\end{tabular}
\caption{Statistics for parallel corpora used in this work. All corpora are parallel between English and the specified `lang`. `\# QA gen' is the number of question-answer pairs generated from each dataset using \paxqa. The bolded GALE dataset has human word alignments, while the others use automated alignments.}
\label{tab:datasets}
\end{table}

\section{Modeling Details}
We release our code, dataset, and documentation.\footnote{\url{https://github.com/manestay/paxqa}} Hyperparameters and other settings are included in those scripts.

Our cross-lingual QA generation method decomposes the task to QG and then MT. We directly use the pretrained question and answer generation model of~\citet{dugan-etal-2022-feasibility}.\footnote{\url{https://huggingface.co/valhalla/t5-base-qa-qg-hl}} We use the lexically constrained MT system of~\citet{wang2022template};\footnote{\url{https://github.com/THUNLP-MT/Template-NMT}} we followed the documentation to train a model for each of the 3 languages. To obtain word alignments, we use \texttt{awesome-align};\footnote{\url{https://github.com/neulab/awesome-align}} we used the provided checkpoint for aligning {zh-en}, while we followed the documentation to train models for {ru-en} and for {ar-en}. 

Our QA models are developed on top of the \texttt{transformers} library. We modify the provided QA training scripts~\footnote{\url{https://github.com/huggingface/transformers/blob/main/examples/pytorch/question-answering/run_qa.py}} for our specific needs.
\section{Additional Results}
\label{sec:add_results}

\paragraph{In-Domain Results}
Results for the \paxqa\hwa\ test set for additional configurations are given in Table~\ref{tab:app_paxqa_results}.
\begin{table*}[ht]
\centering
\small
\setlength{\tabcolsep}{4pt}
\begin{tabular}{@{}lll|ll|ll|lll|l@{}}
\toprule
 & & &
  \multicolumn{2}{c|}{\textbf{non-en $q$}} &
  \multicolumn{2}{c|}{\textbf{en $q$}} &
  \multicolumn{3}{c|}{\textbf{monolingual}} \\ 
Method & Train Data & Train Lang  & en,zh & en,ar & ar,en & zh,en & ar,ar & zh,zh & en,en & avg \\ \midrule
\paxqa\hwa &  lex cons & ar,zh & 88.9  & 91.5  & \textbf{90.4}  & 86.8  & \textbf{90.2}  & 84.1  & \textbf{93.9} & 89.4 \\
\paxqa\hwa & lex cons & zh & 88.2 & 90.4 & 88.0 & 86.7 & 86.4 & 84.0 & 91.9 & 87.9 \\
\paxqa\awa & lex cons & ar,zh,ru & 84.2 & 89.6 & 84.1 & 79.2 & 81.1 & 71.7 & 90.2  & 82.9 \\ \bottomrule
\end{tabular}
\caption{\paxqa\hwa\ test F1 scores for XLM-R models, under various additional configurations. Row 1 is the same as row 3 of Table~\ref{tab:paxqa_results}.
}
\label{tab:app_paxqa_results}
\end{table*}

\paragraph{Generalization Results}
Results for the MLQA test set for additional configurations are given in Table~\ref{tab:app_mlqa_results}. Table~\ref{tab:mlqa_all} reports the averaged F1 and EM scores across all MLQA directions. For our best model configuration (\paxqa\hwa\ lex cons + S), Table~\ref{tab:mlqa_em} gives the individual EM scores, and Table~\ref{tab:mlqa_f1} gives the individual F1 scores.

\begin{table*}[ht]
\centering
\small
\begin{tabular}{@{}lp{1.7cm}l|ll|ll|lll|ll|l@{}} \toprule
    & & &
  \multicolumn{2}{c|}{\textbf{non-en $q$}} &
  \multicolumn{2}{c|}{\textbf{en $q$}} &
  \multicolumn{3}{c|}{\textbf{monolingual}} &
  \multicolumn{2}{c|}{\textbf{non-en xling}} & \\ 
 Method & Train Data & Train Lang & en,zh & en,ar & zh,en         & ar,en & ar,ar & zh,zh & en,en & ar,zh         & zh,ar & avg\\ \midrule

\paxqa\hwa & lex cons + S & ar,zh & \textbf{78.9} & {76.7}          & {70.2}          & \textbf{69.2}          & \textbf{67.0}          & \textbf{71.1}          & \textbf{84.8} & \textbf{64.3} & \textbf{64.9} & 71.9\\
\paxqa\hwa & S + lex cons & ar & 78.0 & \textbf{77.2} & {70.5} & 66.3 & 65.0 & {70.5} & {84.6} & 60.9 & 63.6 & 70.7 \\
\paxqa\hwa & S + lex cons & zh & \textbf{78.9} & 73.7 & 68.7 & {68.3} & {66.7} & 69.2 & 84.2 & {63.9} & 62.4 & 70.7 \\
\paxqa\hwa & S + lex cons + ar$\leftrightarrow$zh pivot & ar,zh & 77.1 & 74.7 & 67.2 & 65.3 & 63.9 & 67.2 & 82.7 & 61.4 & 62.3 & 69.1 \\
\bottomrule
\end{tabular}
\caption{MLQA test F1 scores for \paxqa\hwa\ models, under variable additional configurations. Row 1 is the same as row 5 of Table~\ref{tab:mlqa_results}. Row 4 additionally adds cross-lingual instances between ar and zh, generated through the pivoting strategy described in Section~\ref{sec:ablation}. }
\label{tab:app_mlqa_results}
\end{table*}

\begin{table}[ht]
\centering
\small
\setlength{\tabcolsep}{4pt}
\begin{tabular}{l|ll} \toprule
Method & F1 & EM \\ \midrule
SQuAD (S) & 65.7 & 47.7 \\
\paxqa\hwa\ lex cons + S & 71.1 & 52.9 \\
\citet{riabi-etal-2021-synthetic} & 70.6 & -- \\ \bottomrule
\end{tabular}
\caption{MLQA F1 and EM test scores, averaged across all 49 directions. Note that the \paxqa\hwa\ is zero-shot with respect to 4 of the 7 languages covered by MLQA, while \cite{riabi-etal-2021-synthetic}'s uses all 7 languages.}
\label{tab:mlqa_all}
\end{table}

\begin{table}[ht]
\centering
\small
\setlength{\tabcolsep}{4pt}
\begin{tabular}{l|rrrrrrr} \toprule
c/q & en & es & de & ar & hi & vi & zh \\ \midrule
en & 72.0 & 57.7 & 56.1 & 48.5 & 55.5 & 54.8 & 47.1 \\
es & 67.4 & 56.9 & 56.9 & 46.0 & 51.5 & 52.2 & 42.9 \\
de & 68.6 & 55.3 & 55.2 & 48.3 & 54.8 & 51.9 & 45.3 \\
ar & 63.2 & 51.3 & 51.8 & 46.2 & 48.1 & 50.7 & 43.6 \\
hi & 66.6 & 53.1 & 50.3 & 43.1 & 53.8 & 50.3 & 43.2 \\
vi & 65.4 & 55.6 & 52.4 & 45.4 & 51.4 & 54.3 & 45.7 \\
zh & 65.1 & 53.5 & 51.4 & 43.1 & 50.1 & 50.6 & 47.7 \\ \bottomrule
\end{tabular}
\caption{MLQA EM test scores for each direction, using our best model (\paxqa\hwa\ lex cons + SQuAD).}
\label{tab:mlqa_em}
\end{table}

\begin{table}[ht]
\centering
\small
\setlength{\tabcolsep}{4pt}
\begin{tabular}{l|rrrrrrr} \toprule
c/q & en & es & de & ar & hi & vi & zh \\ \midrule
en & 84.8 & 75.7 & 70.8 & 69.2 & 73.2 & 75.7 & 70.2 \\
es & 81.1 & 75.0 & 72.1 & 66.5 & 69.9 & 72.5 & 67.0 \\
de & 81.3 & 74.0 & 70.0 & 67.8 & 70.6 & 73.2 & 68.6 \\
ar & 76.7 & 68.9 & 65.7 & 67.0 & 65.5 & 71.2 & 64.9 \\
hi & 79.9 & 71.1 & 64.6 & 63.4 & 71.9 & 70.8 & 66.5 \\
vi & 78.4 & 71.7 & 66.1 & 65.5 & 68.7 & 75.2 & 68.4 \\
zh & 78.9 & 71.5 & 67.3 & 64.3 & 68.1 & 72.7 & 71.1 \\ \bottomrule
\end{tabular}
\caption{MLQA F1 test scores for each direction, using our best model (\paxqa\hwa\ lex cons + SQuAD).}
\label{tab:mlqa_f1}
\end{table}

\section{Examples}
\label{sec:examples}
Tables~\ref{tab:sample_zh} and~\ref{tab:sample_ar} show sample \paxqa\hwa\ entries for Chinese and Arabic, respectively. Recall the contexts are drawn from the news domain. For the Chinese sample entry, Table~\ref{tab:translations} compares the question translation of the 3 different MT systems.

\begin{CJK*}{UTF8}{gbsn}
\begin{table*}[ht]
\centering
\small
\begin{tabular}{@{}lp{.85\textwidth}@{}}
\toprule
Field &
  Text \\ \midrule
context$_{en}$ &
  The scientists used a centrifuge from a nuclear weapon manufactured in the former-Soviet Union to obtain high purity silicon, then forged the obtained crystal into the most precise spherosome using hi-tech procedures based on a weight standard of "1kg". At the same time, they used x-ray crystal detector to measure the distance between the spherosome's silicon-28 atoms to determine if the spherosome undergoes obvious atomic changes under certain extreme conditions. In 1889, it was set at a standard one kilogram at the First General Conference of Weights and Measures. \\\midrule
context$_{zh}$ &
  科学家们使用前苏联制造核武器的离心机来提取最高纯度的硅 ， 并把得到的晶体通过高科技手段 ， 按照 “1 公斤 ” 的重量标准打造出这个最精准的圆球体。 同时  ， 利用 X 射线晶体检测器来测量球体硅 － 28 原子之间的空间距离 ， 确定在一些极端条件下该球体不 发生明显的原子变化。 在 1889 年第一届国际计量大会上被定为 1 千克的标准。 \\\midrule
question$_{en}$ &
  What did the scientists use to obtain high purity silicon? \\\midrule
question$_{zh}$ &
  科学家们用什么来获得最高纯度的硅？ \\\midrule
answer$_{en}$ &
  a centrifuge \\\midrule
answer$_{zh}$ &
  离心机 \\ \bottomrule
\end{tabular}
\caption{Sample entry from the \paxqa\textsubscript{HWA} zh-en dataset. The non-English question is translated using lexically-constrained MT.}
\label{tab:sample_zh}
\end{table*}
\end{CJK*}

\begin{table*}[ht]
\centering
\small
\begin{tabular}{@{}lp{.85\textwidth}@{}}
\toprule
Field &
  Text \\ \midrule
context$_{en}$ &
  They cite as evidence the influx of thousands of tourists and visitors to the major international museums to see the best works of classical artists from by-gone ages. They also believe that a great artist can not engage in contemporary, modern and new schools, and be proficient in them, unless he is first proficient in classicism. They recall that Picasso himself, one of the most significant figures to break with classicism, was one of its most proficient exponents in his early career. This also applies to our senior sculpture, Wajih Nahlah, whose seventieth birthday we celebrated yesterday (along with Valentine's day). He himself is a major lover: of the brush, of diligent work, of sublime human and artistic beauty. \\\midrule
context$_{ar}$ & \begin{otherlanguage}{arabic}و يستشهدون ب تدفق آلاف السياح و الزوار على متاحف العالم الكبرى ل مشاهدة روائع الكلاسيكيين من العصور الخوالي. و يرون اكثر: لا يمكن الفنان العظيم ان يخرج الى المدارس العصرية و الجديدة و الحديثة و يبرع في ها, الا اذا كان متمكناً من الكلاسيكية. و يذكرون ان بيكاسو نفس ه, و هو من اكبر من كسروا مع الكلاسيكية, كان من ابرع محترفي ها في مطالع ه الأولى. هذا الكلام ينطبق على كبير نا التشكيلي وجيه نحلة الذي احتفل نا امس ب الذات (مع عيد العشاق) ب عيد ميلاد ه السبعين, و هو نفس ه من كبار العاشقين: ريشةً و عملاً دؤوباً و جمالاً فنياً و انسانياً راقياً. \end{otherlanguage}  \\ \midrule
question$_{en}$ &
  What ideology did Picasso break with in his early career? \\\midrule
question$_{ar}$ &
  \begin{otherlanguage}{arabic} ما هي الإيديولوجية التي انفصل عنها بيكاسو في مطالع ه الأولى؟
  \end{otherlanguage} \\ \midrule
answer$_{en}$ &
  classicism \\\midrule
answer$_{ar}$ &
  \begin{otherlanguage}{arabic} الكلاسيكية
  \end{otherlanguage} \\ \bottomrule
\end{tabular}
\caption{Sample entry from the \paxqa\textsubscript{HWA} ar-en dataset. The non-English question is translated using lexically-constrained MT.}
\label{tab:sample_ar}
\end{table*}

\begin{CJK*}{UTF8}{gbsn}
\begin{table*}[ht]
\centering
\begin{tabular}{@{}ll@{}}
\toprule
System &
  Translated Question \\ \midrule
Vanilla NMT &
  \textcolor{red}{科学家们}用什么来获得高纯度硅？ \\\midrule
Lexically-constrained NMT &
  \textcolor{red}{科学家们}用什么来获得\textcolor{blue}{最高纯度的硅}？ \\\midrule
Google Translate &
  科学家用什么来获得高纯度硅？ \\\midrule
  
\end{tabular}
\caption{Translations from the different systems for the English question ``What did the scientists use to obtain high purity silicon?''. The induced lexical constraints are \textcolor{red}{`the scientists' $\rightarrow$ `科学家们'} and \textcolor{blue}{`high purity silicon' $\rightarrow$ `最高纯度的硅'}, and are highlighted in each translation if they exist. NOTE: in this case, even though only the second translation satisfies all constraints, all 3 translations are grammatically and semantically correct.}
\label{tab:translations}
\end{table*}
\end{CJK*}

\section{QA Generation Evaluation Task}
\label{sec:heval_task}
We sample 2921 QA entries from all QA generations for annotation. These QA entries are generated from randomly sampled articles. The human annotators are drawn from students enrolled in a graduate-level Artificial Intelligence course at an American university. The 129 participants in total were rewarded with extra credit. Annotators are presented the context, the question, and the answer (all in English), and asked the following 3 yes/no questions:
\begin{enumerate}[label=(\roman*).]
    \item Does the question make sense outside of the immediate context?
    \item Is the question relevant and/or interesting?
    \item Is the answer to the question correct?
\end{enumerate}

Because we did not specifically search for bilingual annotators, we evaluated only in English. As a proxy to evaluate answer translations, we propose to back-translate the aligned answers into English. We present this to annotators as an ``Alternate Answer'', and additionally ask:
\begin{enumerate}[label=(\roman*).]
    \setcounter{enumi}{3}
    \item Do ``Answer'' and ``Alternate Answer'' mean the same thing?
\end{enumerate}

\begin{figure*}[ht]
    \centering
    \includegraphics[width=\textwidth]{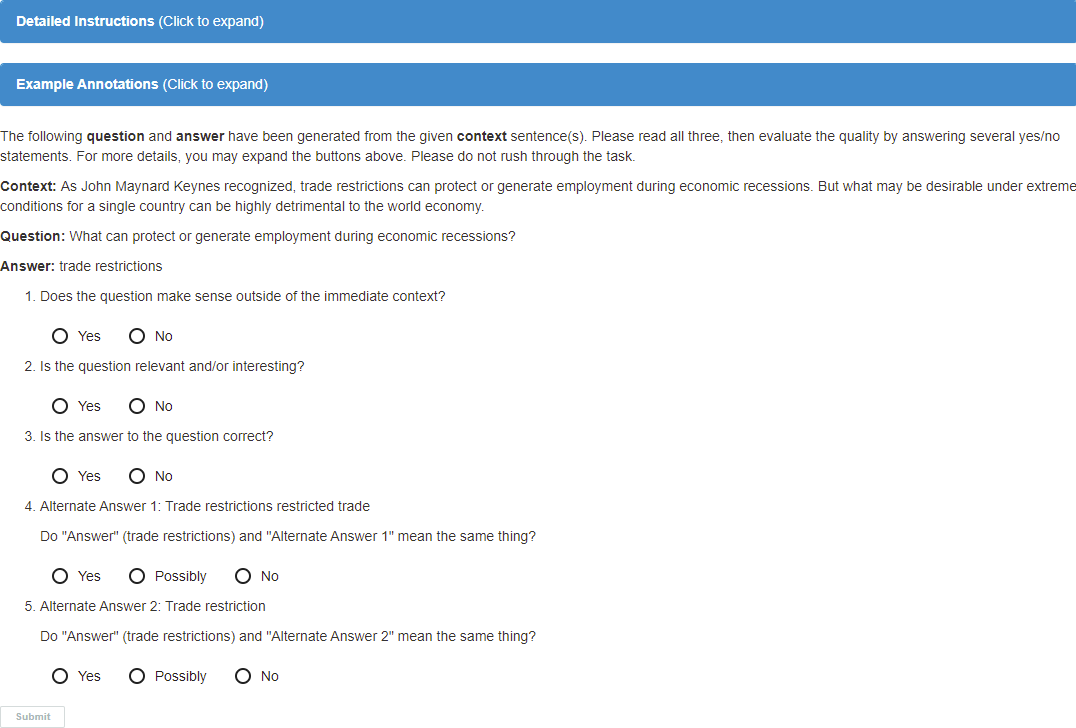}
    \caption{Example QA generation evaluation task presented to human annotators. Note that the task focuses on evaluating the English side of the QA generations. The `Alternate Answer' are the non-English answer spans back-translated to English; we use them as a proxy to evaluate the non-English answers. In this example, the correct answers would be 1. Yes; 2. Yes; 3. Yes; 4. Possibly; 5. Yes.}
    \label{fig:mturk}
\end{figure*}

The annotation interface is shown in Figure~\ref{fig:mturk}. We collect 3 annotations per task, and assign the majority label as the gold label. 

We evaluate inter-rater reliability using averaged pair-wise Cohen's kappa $\kappa$: (i) 0.18, (ii) 0.18, (iii) 0.41, (iv) 0.51. The $\kappa$ scores for (i) and (ii) are especially low, which indicates that workers had very subjective understandings of interpretability and relevance. This is likely because we did not train workers, and merely provided them with the instructions. $\kappa$ for (iii) and (iv) indicate moderate agreement. 

We define `high-quality' QA entries with the following criterion:

$\big($(i) is `Yes' OR (ii) is `Yes'$\big)$ AND $\big($(iii) is `Yes'$\big)$  AND $\big($(iv) is `Yes'$\big)$ 

From the 2921 annotations, we filter to 1724 (59.0\%) high-quality QA entries, which we then assign to the \paxqa\hwa\ validation and test sets.

In other words, for the English QG model used in this work, ~59.0\% of generations were deemed acceptable by human annotators. As our methodology supports drop-in replacements, we suspect that using better QG models will improve QG quality, and likely downstream QA performance.

\section{GenBench Evaluation}
\begin{figure*}[ht]
    \centering
    \includegraphics[width=0.8\textwidth]{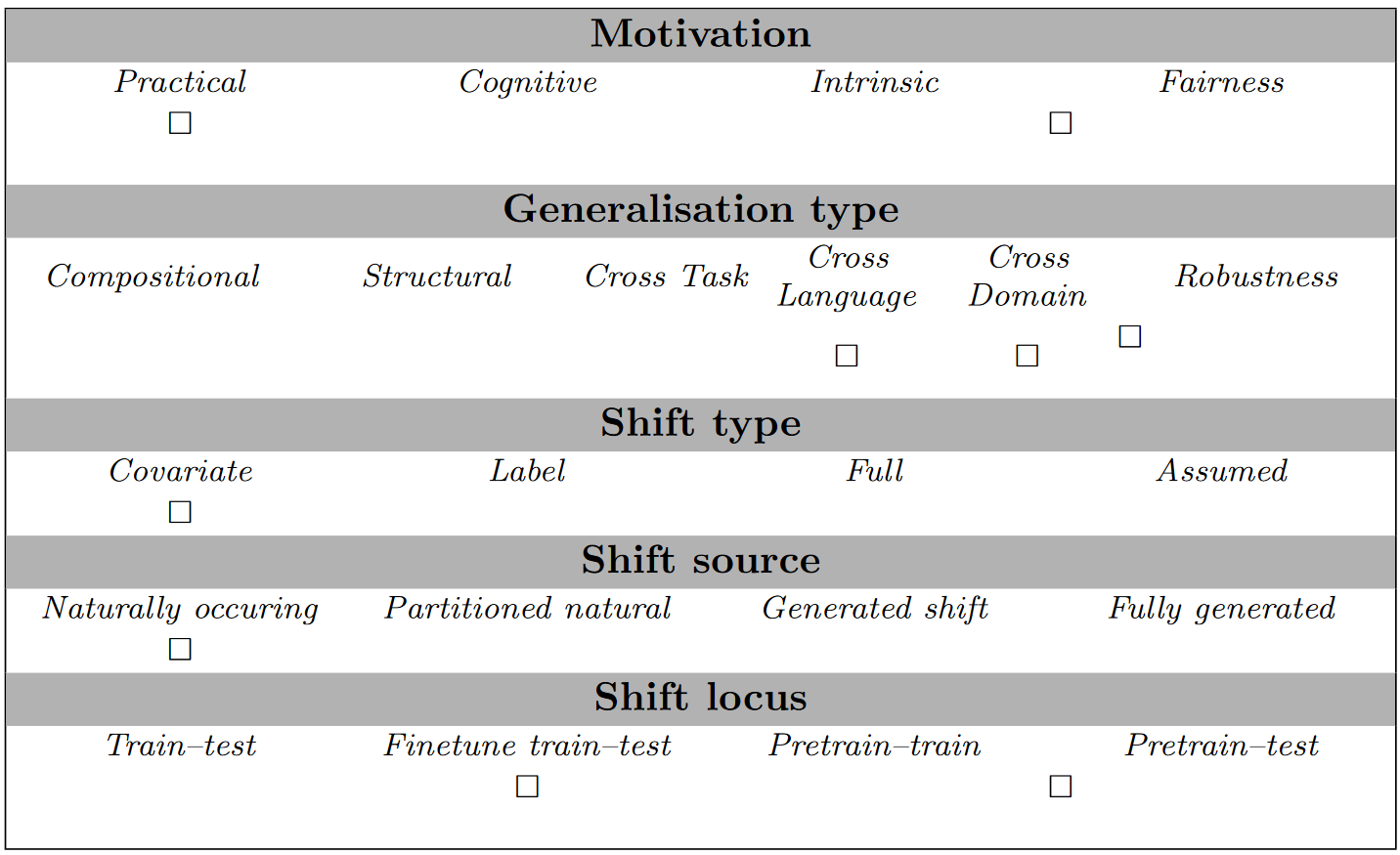}
    \caption{GenBench evaluation card for the experiments studied in this work.}
    \label{fig:eval_card}
\end{figure*}

Our work aims to be included as part of the GenBench initiative, and the evaluation card is shown in Figure~\ref{fig:eval_card}. We now give the rationales for the GenBench taxonomy axes.

\paragraph{Motivation:} our study is \textit{practical} in that it studies how we can generate synthetic data to fine-tune cross-lingual QA models, and considers \textit{fairness} in studying multiple languages.

\paragraph{Shift Type:} our study considers \textit{covariate} shift, since the inputs change by languages and textual domains. Note that for the extractive QA task considered, given the input, the distribution of labels is the same, i.e., always a span from the input.

\paragraph{Shift source:} the shifts are \textit{naturally occurring}, with the same logic that the inputs change by languages and textual domains.

\paragraph{Shift locus:} we mainly studied \textit{finetune train-test} for \paxqa-finetuned models, but also \textit{pretrain-test} for the zero-shot baseline.

\paragraph{Generalisation type:} our experiments touch upon three types, which we further discuss below.
\begin{itemize}
    \item \textbf{Cross language}
We trained models on QA pairs generated from the news domain (\paxqa), then tested on QA datasets from the Wikipedia domain (MLQA, \textsc{XorQA}. For example, see Table~\ref{tab:mlqa_results}.
    \item  \textbf{Cross domain}
We trained models on cross-lingual QA pairs for several languages, then tested on other language directions. For example, see Tables~\ref{tab:xorqa_results} and~\ref{tab:mlqa_all}.
    \item \textbf{Robustness} 
We trained models on QA pairs generated while using noisy word alignments. For example, see rows 8 and 9 of Table~\ref{tab:mlqa_results}.
\end{itemize}

